\documentclass[letter, 10pt, conference]{ieeeconf}
\IEEEoverridecommandlockouts

\usepackage{cite}

\usepackage{amsmath,amssymb,amsfonts}
\usepackage{algorithmic}
\usepackage{graphicx}
\usepackage{subcaption}
\usepackage{textcomp}
\usepackage{xcolor}
\usepackage{bm}
\usepackage{booktabs}
\usepackage{tikz}
\usepackage{hyperref}

\def\BibTeX{{\rm B\kern-.05em{\sc i\kern-.025em b}\kern-.08em
    T\kern-.1667em\lower.7ex\hbox{E}\kern-.125emX}}
\usetikzlibrary{arrows,decorations.pathmorphing,backgrounds,fit,positioning,calc,shapes}

\begin{document}

\title{Interactive Disambiguation for Behavior Tree Execution
\thanks{
\textbf{This work has been submitted to the IEEE for possible publication. Copyright may be transferred without notice, after which this version may no longer be accessible.}
This project is financially supported by the Swedish Foundation for Strategic Research.%
It is also partially funded by grants from the Swedish Research Council (2017-05189), the Swedish Foundation for Strategic Research (SSF FFL18-0199). The authors gratefully acknowledge this support.}
}

\author{\authorblockN{
Matteo Iovino$^{a,b}$,
Fethiye Irmak Do\u{g}an$^{a}$,
Iolanda Leite$^{a}$,
Christian Smith$^{a}$}
\thanks{$^{a}$Division of Robotics, Perception and Learning, KTH - Royal Institute of Technology, Stockholm, Sweden}
\thanks{$^{b}$ABB Corporate Research, Västerås, Sweden}
}

\maketitle

\begin{abstract}
In recent years, robots are used in an increasing variety of tasks, especially by small- and medium  sized enterprises. These tasks are usually fast-changing, they have a collaborative scenario and happen in unpredictable environments with possible ambiguities. It is important to have methods capable of generating robot programs easily, that are made as general as possible by handling uncertainties. We present a system that integrates a method to learn Behavior Trees (BTs) from demonstration for pick and place tasks, with a framework that uses verbal interaction to ask follow-up clarification questions to resolve ambiguities. During the execution of a task, the system asks for user input when there is need to disambiguate an object in the scene, when the targets of the task are objects of a same type that are present in multiple instances. The integrated system is demonstrated on different scenarios of a pick and place task, with increasing level of ambiguities. The code used for this paper is made publicly available\footnote{\url{https://github.com/matiov/disambiguate-BT-execution}}.
\end{abstract}


\section{Introduction}


\begin{figure}[tbp]
    \centering
    \includegraphics[width=.9\linewidth]{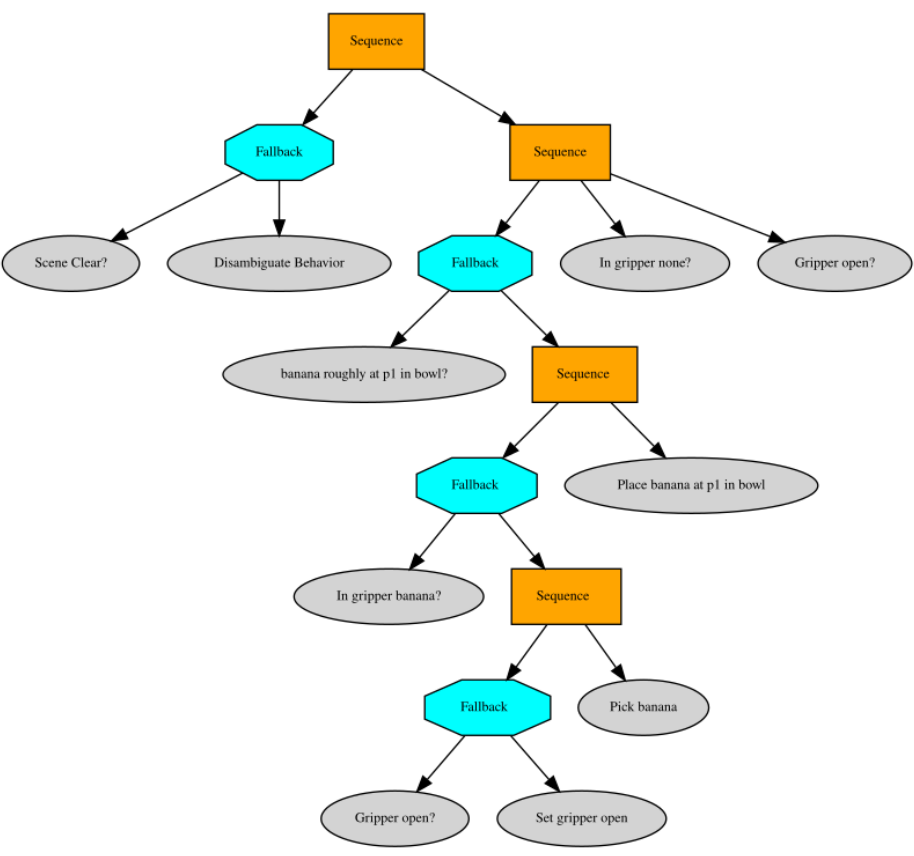}
    \caption{Behavior Tree with the disambiguation subtree, to solve a manipulation task where a banana has to be placed in a bowl.}
    \label{fig:task_BT}
\end{figure}

Modern industrial robotic applications are becoming more dynamic and complex. Robots are tasked to solve a broader range of problems and to operate in frequently changing environments shared with humans. This high variety of tasks requires easy and rapid generation of new robot programs, with the ability to react to changes in the environment. These requirements can be met by using learning algorithms to synthesize control policies that are reactive, human-readable, and modular. Behavior Trees (BTs) have proven to be a good policy representation for many robotic applications~\cite{colledanchise_behavior_2018} and numerous works propose methods to automatically synthesize them~\cite{iovino_survey_2020}.

Additional challenges arise if the robot relies on vision capabilities. Markers allow to label items in the scene univocally, but they are not suited for industrial robotic applications or when it is desired to manipulate common objects or tools. Common object recognition algorithms can detect objects in the scene but may not be able to distinguish between objects belonging to the same category or class. 
A possible use-case scenario could be one with a robot tasked to prepare a kit-box to deliver to a human operator for an assembly operation. The robot might have access to a shelf where common tools are stored, like screwdrivers and hammers. Supposing that the object detection algorithm is able to tell a screwdriver from a hammer, there are still ambiguities on which type of screwdriver is suitable for the subsequent assembly task. These ambiguities could be solved at run-time by letting the robot verbally interact with the human. The robot could ask efficient follow-up questions by using the available information and by making reasonable suggestions to the collaborator while asking for clarification. To this extent, referring expressions, i.e. phrases that describe objects with their distinguishing features, can be used.

In this paper, we propose to integrate the interactive disambiguation framework from~\cite{dogan_followup_2022} with a method for learning BTs from demonstration~\cite{gustavsson_combining_2021}, to solve ambiguities that might arise during execution of pick and place tasks. We use Learning from Demonstration (LfD) methods as they enable intuitive generation of robot programs, with small effort required by the user. We make the additional assumption that the BT is learned in a non-ambiguous environment, but that ambiguities might occur during the execution of the task. Therefore, our contribution is an integrated system that learns BTs from demonstration that allow the robot to solve ambiguities at execution time, through verbal interaction with a human user. We call a task or a scene \emph{ambiguous} if the robot has to interact with objects of a category that are present in multiple instances. If there is only a single item of the target object category present, then the scene is \emph{non-ambiguous}. The task is disambiguated with clarifying questions, where one of the objects is pointed at unequivocally as the target for the task. In the reminder of this paper we present the background on BTs and LfD and the related work in Section~\ref{sec:work}, a detailed description of the integrated system and the underlying methods in Section~\ref{sec:method}, to conclude with a set of experiments designed to validate the proposed system in Section~\ref{sec:experiments}.

\section{Background and Related Work} \label{sec:work}

This section provides a background on Behavior Trees and Learning from Demonstration and summarizes related work, highlighting the uniqueness of our systems.

\subsection{Behavior Trees}

Behavior Trees are task switching policy representations, historically conceived as an alternative architecture to Finite State Machines (FSM)~\cite{colledanchise_behavior_2018}. They have explicit support for task hierarchy, action sequencing and reactivity, and improve on FSMs especially in terms of modularity and reusability~\cite{iovino_survey_2020}. 
The internal nodes of a BT are called \emph{control nodes}, while leaves are called \emph{execution nodes} or \emph{behaviors} (polygons and ovals, respectively, in Figure~\ref{fig:task_BT}). At run-time, the BT is \textit{ticked} from the root down the tree at a specified frequency. The most common control nodes are \emph{Sequence} and \emph{Fallback} (or \emph{Selector}). The former execute their children in a sequence, returning once all succeed or one fails. The latter also execute their children in a sequence, returning when one succeeds or all fail. Execution nodes execute a behavior when ticked and return one of the status signals \textit{Running}, \textit{Success} or \textit{Failure}. They are of type Action nodes or Condition nodes, the latter immediately returning \textit{Success} or \textit{Failure} as they encode status checks and sensory feedback. BTs can functionally be compared to decision trees, but the \textit{Running} state allows BTs to execute actions for longer than one tick. The \textit{Running} state is also crucial for reactivity, allowing other actions to preempt non-finished ones. As discussed in~\cite{gustavsson_combining_2021}, reactivity is inherently gained when representing a policy learned from demonstration as a BT. Further detail on BTs is found in e.g.~\cite{colledanchise_behavior_2018}.

\subsection{Learning from Demonstration}

Learning from Demonstration (LfD - also known as Programming by Demonstration or Imitation Learning) defines those methods that allow to generate robot programs from human demonstrations~\cite{ravichandar_recent_2020}, thus lowering both the requirements on the programming skills and time needed to write robot programs. A LfD method defines how tasks are demonstrated and how the robot policies are represented and learned. Demonstrations can mainly be of three types: \textit{kinesthetic teaching}, where the user physically moves the robot, \textit{teleoperation}, where a robot is controlled through an external device and \textit{passive observation}, where the robot or the human are endowed with tracking systems and the demonstrator's body motion is recorded.
Each demonstration method has pros and cons, most importantly about correspondence problem, i.e. the mapping between a motion performed by a human teacher and the one executed by the robot.
In Kinesthetic teaching, used in this paper, the motion is directly recorded in the robot task (or joint) space, so it does not suffer form the correspondence problem. This method is intuitive and only minimal instruction is required for the user. The main drawback is the limit in the number of robot degrees of freedom and mass that a human can move to perform a demonstration.

\subsection{Related Work}


Despite the increased interest in BTs in the AI community in recent years, only a few methods exist that are capable of automatically generating the structure of a BT solving a robotic task. Some examples are automatic planners~\cite{rovida_extended_2017,colledanchise_synthesis_2017, colledanchise_towards_2019}, genetic programming~\cite{iovino_learning_2021, styrud_combining_2022} and learning from demonstration~\cite{french_learning_2019, gustavsson_combining_2021}. 
Methods in~\cite{french_learning_2019} and~\cite{sagredo-olivenza_trained_2019} learn a mapping from state space to action space as a Decision Tree (DT), which is converted into a BT, since BTs generalize DTs \cite{colledanchise_how_2017}. In~\cite{robertson_building_2015}, on the other hand, demonstrations are recorded in the form of a sequence of actions and directly stored in a BT. In both~\cite{sagredo-olivenza_trained_2019} and~\cite{robertson_building_2015} the target application is controlling game AI, although trees learned in the former are used as guide to assist game designers. The method applied in~\cite{robertson_building_2015} aims to learn a strategy to play StarCraft, but the outcome is a very large and hard to read BT ($>50.000$ nodes) whose structure limits the reactivity. The BT learned in~\cite{french_learning_2019} is implemented on a mobile manipulator performing a house cleaning task. At the learning stage, the whole action space and state space are encoded in the tree, affecting its size. The reference frames for all actions are hard-coded, thus requiring additional effort from the programmer and reducing both reusability and the ability of the method to generalize. The method proposed in~\cite{knaust_guided_2021} uses human demonstrations to learn robot skills that are encoded as a BT. In particular, what is learned is an action combining Probabilistic Movement Primitives (ProMPs), point-to-point and gripper motions. ProMPs are low dimensional representation for trajectory distributions that can exploit the variance in the demonstrations.  
In contrast to~\cite{knaust_guided_2021} we do not learn trajectories from the demonstrations, only the end point of the action, and we let a motion planner compute the optimal trajectory between start and end points. We learn the whole structure of the BT for the demonstrated task, while in~\cite{knaust_guided_2021} the structure is fixed.

Early attempts on implementing HRI in BTs were made in~\cite{hu_semi-autonomous_2015, hu_semi-autonomous_2018}, where a surgical robot is controlled by BT. The robot execution halts before performing the tumor ablation and user input is required to decide one of the two possible ablation paths to proceed. Authors in~\cite{fusaro_human-aware_2021, fusaro_integrated_2021} modify the formalism of the BT introducing different cost types for every action performed by a robot. At execution time the control node selects the action with the lowest cost. In a collaborative industrial scenario, an operator can communicate their intentions to the robot, that keeps track of the actions executed by the human and focuses on the remaining ones. Finally,~\cite{coronado_towards_2021} proposes \textit{OpenRIZE}, a GUI that allows users to program robot interaction behaviors for HRI applications. The execution layer builds on BTs. In all these methods, the interaction with the robot is realized through a GUI.
The work by Suddrey et al.~\cite{suddrey_learning_2021} is the most similar to ours. The method maps verbal instructions to BTs with natural language processing. First, the instruction is parsed and a register of previously learned BTs is accessed to find matches. Otherwise, a new BT is learned from scratch. During a demonstration, the robot can ask questions to disambiguate the scene by using object properties (e.g. color). In our work, intrinsic object properties are replaced with spatial relations, and disambiguation is done during the execution step. 
As the system in~\cite{suddrey_learning_2021} has no frame inference, the robot cannot solve more complicated assembly tasks.

For treating ambiguities, previous studies have focused on asking follow-up questions for various tasks~\cite{romanroman:EMNLP-Findings20,amiri2019augmenting}, and interactive clarification has been suggested to enhance human-robot interaction while disambiguating described objects~\cite{hatori2018interactively, shridhar2018interactive,dogan_followup_2022}. An interactive system has been presented where the robot proposes the possible target objects and asks an operator about which one to pick up when the initial object description is ambiguous~\cite{hatori2018interactively}. In another study, Shridhar et al.~\cite{shridhar2018interactive, shridhar2020ingress} have suggested a method to decode the users' object descriptions using \emph{grounding by generation}. This enables their system to generate object-specific follow-up questions to resolve ambiguities. Further, Amiri et al.~\cite{amiri2019augmenting} have proposed to augment the robot's knowledge base with different entities by asking follow-up questions such as \textit{`Where should I deliver the coffee?'}. Although there have been promising attempts to disambiguate described objects, previous studies mostly assumed that the target object candidates are given or could be detected or localized by existing methods. 
To address these limitations, Dogan et al.~\cite{dogan_followup_2022} presented an autonomous system to generate follow-up clarifications using the known objects in the environment, and the recognized part of the user request without putting the previous constraints on object categories. They have shown that when there are uncertainties, the robot can find the described object more often with fewer conversational turns when it asks for follow-up clarifications instead of another description of the same object. To generate the clarifications, they have used the spatial relations between objects, and we employ their approach in our system, as detailed in Section~\ref{disambiguation}.
In their system, they have a component to identify whether the request is ambiguous, but we adapt this system by limiting the interaction with the human only to ambiguous cases, letting the robot resolve the disambiguated task as previously demonstrated.

To summarize, our proposed system that integrates BTs learned from human demonstration with a method that uses follow-up questions to solve disambiguities in the scene is able to solve some of the shortcomings in the state of the art. For what concerns BT learning, we learn a tree that is robust to configuration changes in the environment because the method is able to infer the relevant components of a task, as will be explained in the next section. As per the disambiguation framework, we limit the interaction with the user to the necessary situations to solve the task: the human doesn't need to query the robot with an object to manipulate, as the underlying task has already been shown during a demonstration.

\section{Proposed Integrated System} \label{sec:method}




\begin{figure}[tbp]
    \centering
    \includegraphics[width=.5\linewidth, height=3.5cm]{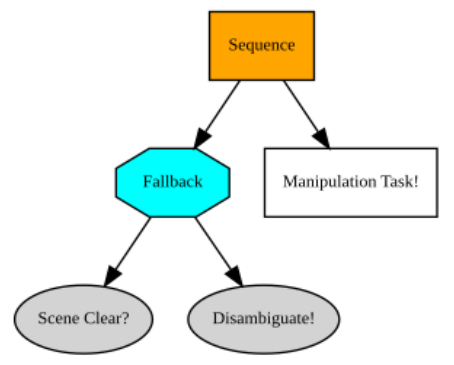}
    \caption{High-Level BT generated by the system: the manipulation BT that is learned from demonstration is attached in sequence to a subtree for disambiguating the scene.}
    \label{fig:combinedBT}
\end{figure}

\begin{figure}[tbp]
    \centering
    \includegraphics[width=.6\linewidth]{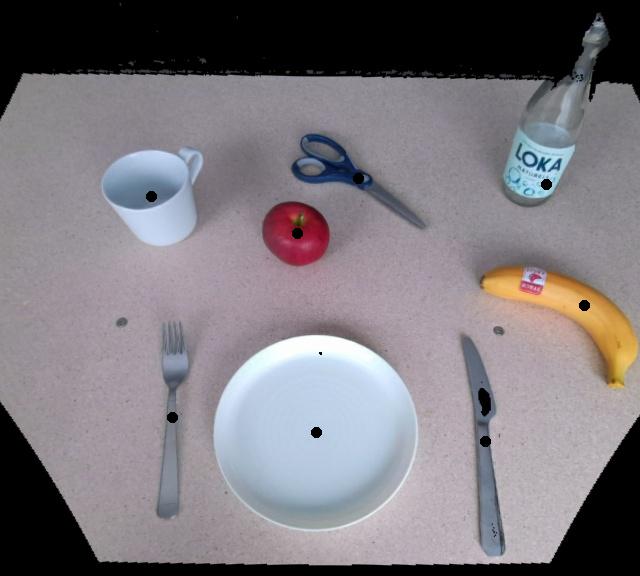}
    \caption{The black dots are the grasping points in the image, found with object-specific heuristics to the depth image.}
    \label{fig:heuristic}
\end{figure}

\begin{figure*}
     \centering
     \begin{subfigure}[b]{0.9\textwidth}
         \centering
         \includegraphics[width=\textwidth]{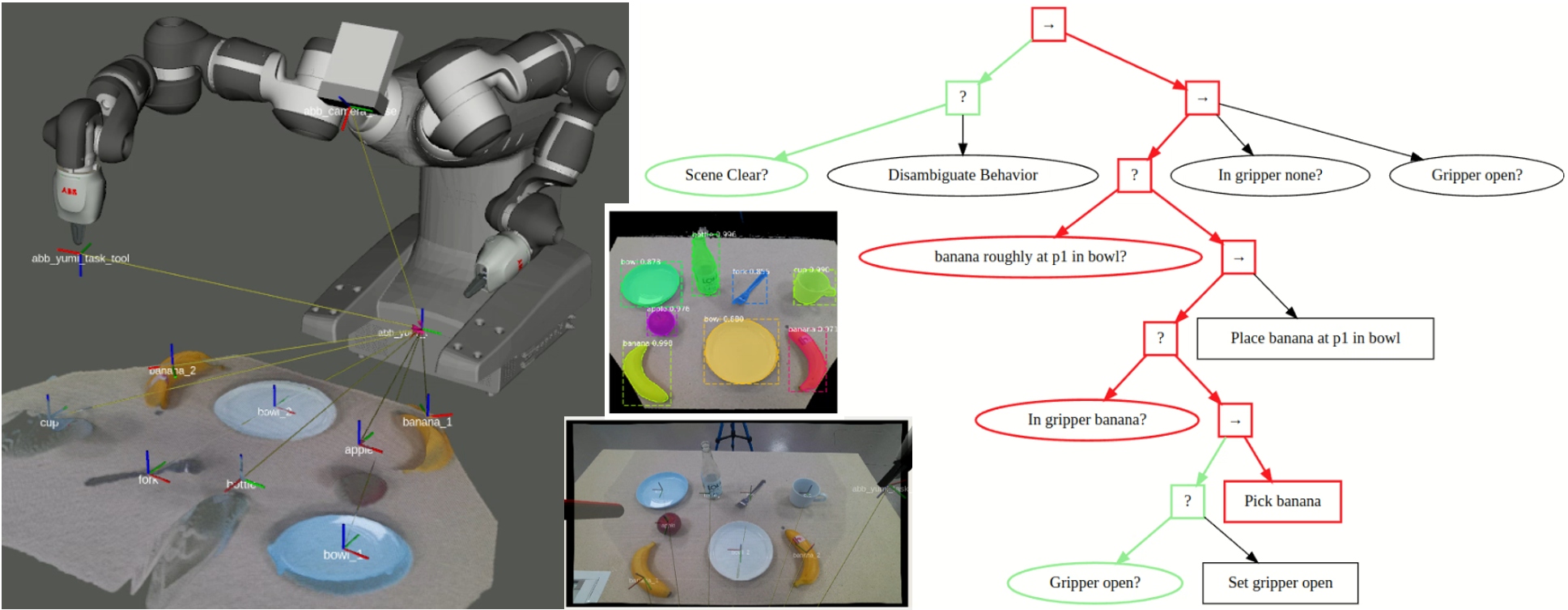}
         \caption{\textbf{Initialization.} The \emph{`Scene Clear?'} condition is initialized as \texttt{True}. The BT ticks the manipulation subtree, but the \emph{Pick} action and the \emph{`Object Roughly At?'} condition turn false because both the \emph{bowl} and the \emph{banana} are ambiguous. The next time the BT is ticked, the \emph{`Scene Clear?'} will turn to \texttt{False} and the \emph{`Disambiguate Behavior'} will be executed, first with query \textit{bowl} then with query \textit{banana}.}
         \label{fig:step1}
     \end{subfigure}
     \hfill
     \begin{subfigure}[b]{0.9\textwidth}
         \centering
         \includegraphics[width=\textwidth]{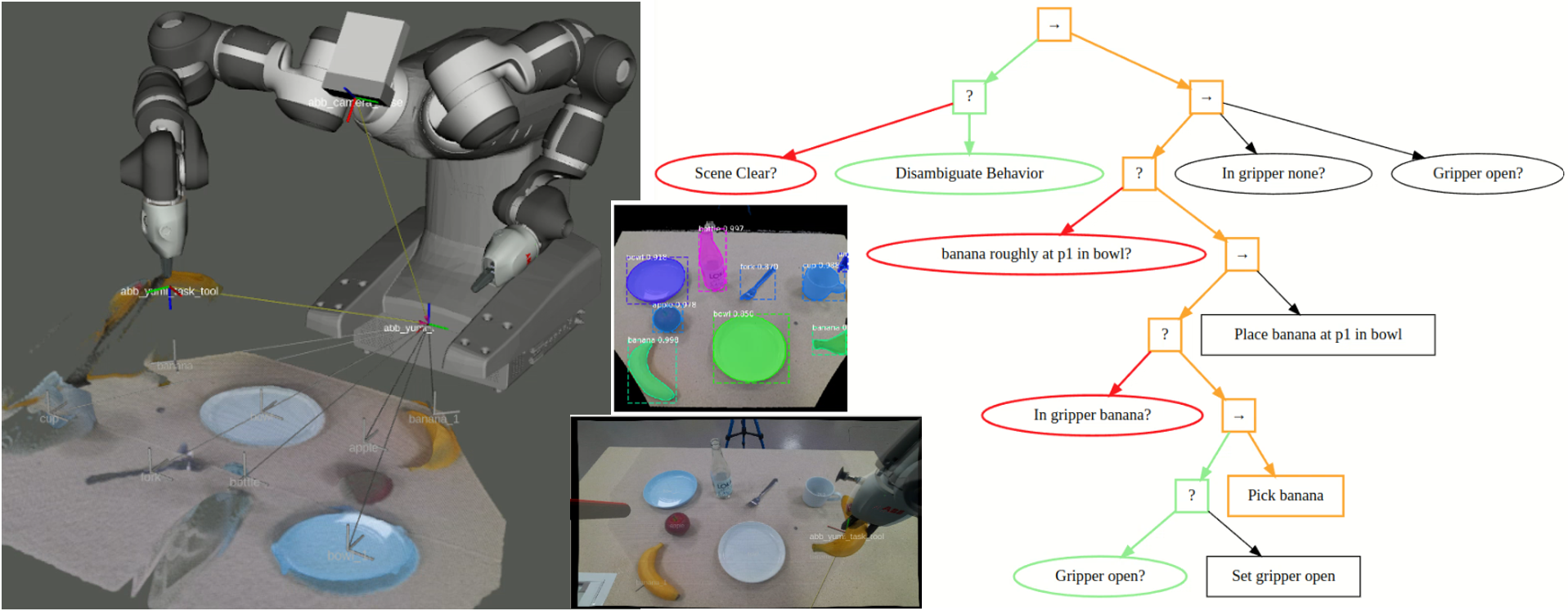}
         \caption{\textbf{Picking.} Now that the objects are disambiguated, the name of the respective reference frame changes accordingly and the manipulation task can be executed. The robot is picking the target item \emph{banana}. During the manipulation, the detection algorithm is timed out and the reference frames are not updated.}
         \label{fig:step2}
     \end{subfigure}
     \hfill
     \begin{subfigure}[b]{0.9\textwidth}
         \centering
         \includegraphics[width=\textwidth]{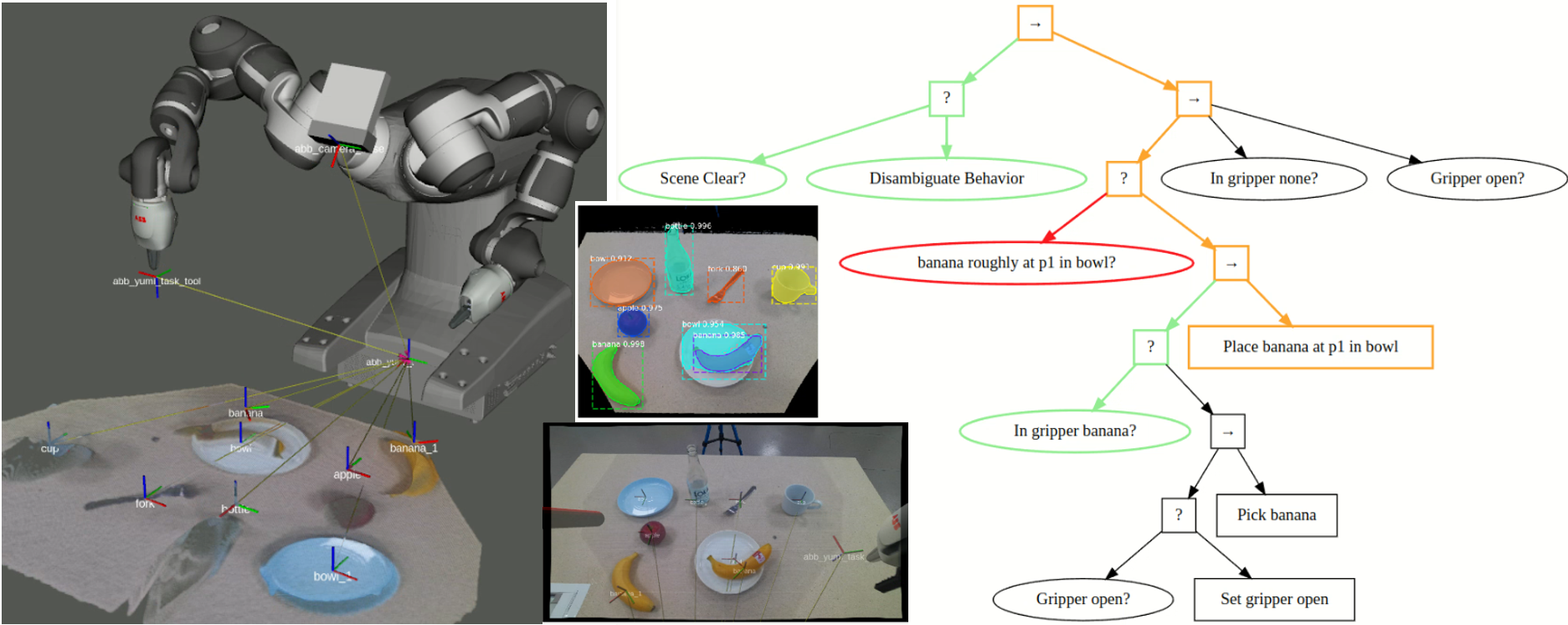}
         \caption{\textbf{Placing.} The robot is placing the target item \emph{banana} in the reference frame of the \emph{bowl}. When the detection runs again, the system is able to update correctly the frames of the disambiguated objects. The next time the BT is ticked, also the \emph{`Object Roughly At?'} condition will return \textit{Success} and the task will be concluded.}
         \label{fig:step3}
     \end{subfigure}
        \caption{Execution pipeline of the system with two ambiguous items. In the Behavior Tree, \emph{`$\rightarrow$'} represents a \emph{Sequence} node; \emph{`$?$'} represents a \emph{Fallback} node; green, yellow and red represent the return statuses \emph{Success}, \emph{Running} and \emph{Failure} respectively.}
        \label{fig:experimentE}
\end{figure*}

The proposed system to disambiguate robot manipulation tasks at execution time builds on an integration of the method to learn Behavior Trees from Demonstration proposed in~\cite{gustavsson_combining_2021} and the method to disambiguate scenes by using verbal interaction proposed in~\cite{dogan_followup_2022}. The system combines both methods to build BTs under the hypothesis that the demonstrated task is not ambiguous, but that the scene where the task is executed may be.
In this scenario the same task is demonstrated three times, to allow the system to generalize the task, e.g. infer in which reference frames the actions are executed.All three task demonstrations have the same objects in the environment but their initial position is changed every time.  Multiple demonstrations result in more robust programs. Otherwise, the robot would just directly copy the human actions, but be unable to adapt to any changes in the objects' initial states.. 
Then, the learned tree is automatically extended to include the disambiguation subtree (rooted with a \textit{Fallback} node in Figure~\ref{fig:combinedBT}).

Finally, the tree is run to solve the demonstrated task in a new scenario, that can have none or multiple ambiguities. In the ambiguous situations, every object that is present with multiple instances is attached to a reference frame with name \textit{object\_id}: for example, if there are two bananas, one is attached to a frame with name \textit{banana\_1} while the other to a frame with name \textit{banana\_2} (as in Figure~\ref{fig:step1}). Since the tree is learned in non-ambiguous environments, the targets for a manipulation task would be identified by a frame with the simple name \textit{object}. The main focus of this paper is not vision based grasping, so here we computed the frames using object specific heuristics. To this extent, the Mask-RCNN detection algorithm~\cite{matterport_maskrcnn_2017} is used to obtain objects bounding boxes and masks, which are then used to compute grasping points, first in the image space (the black dots in Figure~\ref{fig:heuristic}) and then in the world space, using the camera matrix for projection. A reference frame is attached to the point obtained in this way (righ-hand sides of Figure~\ref{fig:experimentE}).

During execution, the condition \textit{`Scene Clear?'} is initialized as \texttt{True}, so that the robot attempts to solve the manipulation task (Figure~\ref{fig:step1}). If the task fails, the condition turns to \texttt{False} and the disambiguation method is run. If the method completes successfully, the disambiguated object's reference frame is assigned the correct name (as in Figure~\ref{fig:step3} for the items \textit{banana} and \textit{bowl}) and the manipulation task is attempted again. Note that hard-coding the frame names in the demonstration would cause a loss in generality and reactivity. To keep track of the relation between frames and objects, a dictionary of detected items is updated every time the detection is run, making it possible to rename the frames dynamically if new objects appear in the scene or disappear from it. The detection is halted during the manipulation action, to discard the noise caused by the robot arm passing in the field of view of the camera (in Figure~\ref{fig:step2} the reference frames are disappearing as the robot is attempting picking and then placing). Afterwards information about the detected objects is updated, with the assumption that only one object can move in the halted interval. 
The interaction between the disambiguation and manipulation subtrees is realized by accessing variables stored in a blackboard.

\subsection{Learning BTs from Demonstration}

\begin{figure}[tbp]
    \centering
    \includegraphics[width=\linewidth]{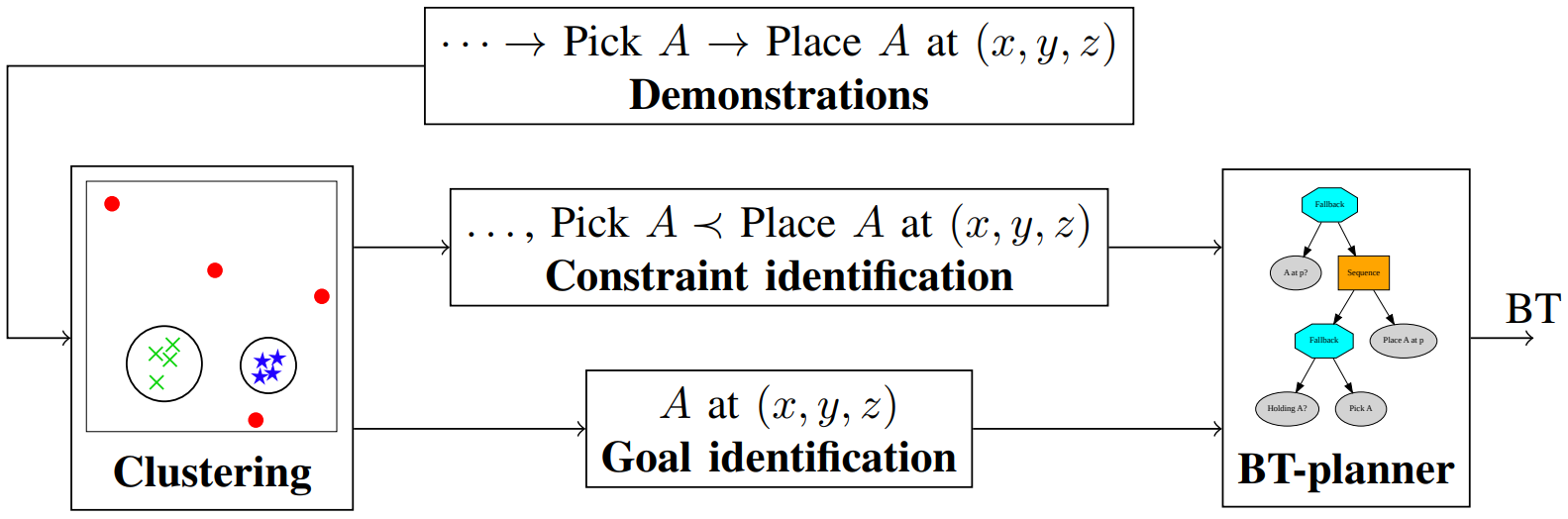}
    \caption{Outline of the Learning from Demonstration method.}
    \label{fig:LfDoutline}
\end{figure}

\begin{table}[tbp]
\centering
\caption{Available actions with their description.}
\label{tab:actions}
\begin{tabular}{ll}
\toprule
\textbf{Action} & \textbf{Description} \\ \midrule
\textsc{Pick} & Close the grippers around the target object. \\
 & \textbf{Pre-condition:} gripper open. \\
 & \textbf{Post-condition}: gripper close and object picked.\\
 & \textbf{Parameters:} object to pick. \\
 & \textbf{BT action:} `pick \texttt{object}?'. \\
 & \textbf{BT condition:} `in gripper \texttt{object}?'. \\

\textsc{Place} & Open the gripper. Intended as a precise place. \\
 & Tolerance for the place pose: sphere of radius $3$~cm. \\
 & \textbf{Pre-condition:} gripper close and item picked. \\
 & \textbf{Post-condition}: gripper open.\\
 & \textbf{Parameters:} object to place / target pose / target frame.\\
 & \textbf{BT action:} `place \texttt{object} at \texttt{pose} in \texttt{frame}?'. \\
 & \textbf{BT condition:} `\texttt{object} at \texttt{pose} in \texttt{frame}?'. \\

\textsc{Drop} & Open the gripper. Intended as a rough place. \\
  & Tolerance for the place pose: cylinder of radius $10$~cm \\
  & \textbf{Pre-condition:} gripper close and item picked. \\
  & \textbf{Post-condition}: gripper open.\\
  & \textbf{Parameters:} object to place / target pose / target frame.\\
  & \textbf{BT action:} `place \texttt{object} at \texttt{pose} in \texttt{frame}?'. \\
  & \textbf{BT condition:} `\texttt{object} roughly at \texttt{pose} in \texttt{frame}?'. \\
\bottomrule
\end{tabular}
\end{table}

The algorithm proposed in~\cite{gustavsson_combining_2021}  learns BTs from demonstrations in four steps, as shown in Figure~\ref{fig:LfDoutline}. The demonstrations are performed with kinesthetic teaching. During the demonstration the robot is set to \textit{LeadThrough Mode}: the robot arm has the motor brakes switched off and is gravity balanced, to allow the human to move it around effortlessly. Once the desired position in Cartesian space is reached, the human can select one of the actions described in Table~\ref{tab:actions}. A \textit{Pick} action will close the robot grippers around the target object and a \textit{`Place'} or \textit{`Drop'} action will open the grippers, releasing the object. For all actions, the pose of the end-effector is recorded as the target pose for that action, in all possible reference frames. All actions are defined together with their pre- and post-conditions.
The difference between \textit{`Place'} and \textit{`Drop'} lies in the accuracy with which the object is placed. In \textit{`Place'}, the tolerance is within a sphere of radius $3$~cm centered in the goal position of the action, while in \textit{`Drop'} a cylinder of radius $10$~cm is used. Thus a \textit{`Drop'} action is chosen when the object goal pose is not relevant, e.g. when dropping objects in a trash bin.
Then, different demonstrations of the same task are clustered together to infer the reference frame of each action, the task constraints and goal conditions.

\begin{figure}[tbp]
     \centering
     \begin{subfigure}[b]{0.45\textwidth}
         \centering
         \includegraphics[width=\textwidth]{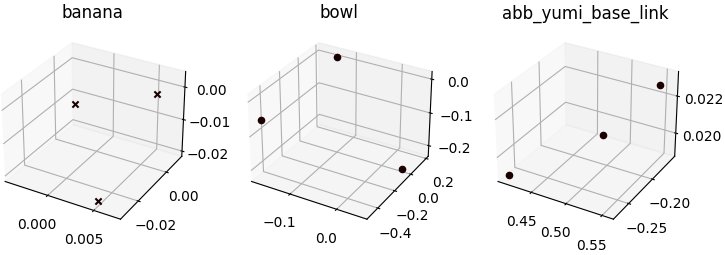}
         \caption{Clustering step for the PICK action: the action happens in the reference frame of the banana.}
         \label{fig:cluster_pick}
     \end{subfigure}
     \hfill
     \begin{subfigure}[b]{0.45\textwidth}
         \centering
         \includegraphics[width=\textwidth]{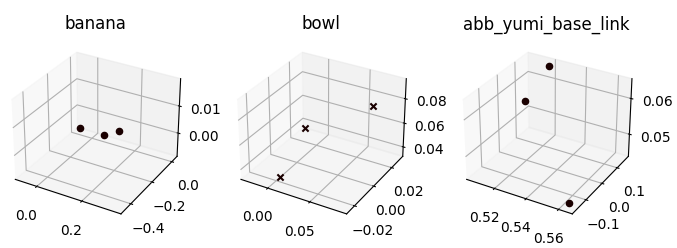}
         \caption{Clustering step for the DROP action: the action happens in the reference frame of the bowl (for obvious reasons, the banana is disregarded).}
         \label{fig:cluster_place}
     \end{subfigure}
        \caption{Frame inference. During the learning process, for each action the pose of the end effector is recorded in all relevant reference frames. Then, the samples are clustered and the frame that forms the smallest cluster is taken as reference frame for the action.}
        \label{fig:clustering}
\end{figure}

Frame inference is realized with an unsupervised clustering algorithm. Similar actions that are executed in the reference frame $F$ will lie close together and form clusters, so candidate actions' target positions are represented in each of the candidate frames. By default, actions are set to be executed in the robot base frame.
As an example, to teach the robot how to pick a banana and drop it in a bowl (represented by the BT in Figure~\ref{fig:task_BT}), the same demonstration is repeated (at least) three times (three is the sample threshold for the clustering algorithm), with different starting configurations for the objects. For this example, the pose of the end effector is recorded in the reference frame of the robot, the banana, and the bowl (Figure~\ref{fig:clustering}). 

In the final step a planner adapted from~\cite{colledanchise_towards_2019} builds the BT using \emph{backchaining}: starting from the goal, pre-conditions are iteratively expanded with actions that achieve them - those actions that have that particular condition as their post-conditions. Then, those actions' unmet pre-conditions are expanded in the same way. For example, for the BT showed in Figure~\ref{fig:task_BT}, using as reference Table~\ref{tab:actions}. A \emph{`Drop'} action has the post-condition of the target object to be placed roughly in a position in a specific reference frame, thus the condition \emph{`Object Roughly At?'} is inserted automatically under a \textit{Fallback} node. If the \emph{`Drop'} action fails during planning, it is expanded with its pre-condition, e.g. to place an object you first have to pick it. Thus, a subtree with the \emph{`Pick'} action is added and its pre- and post-conditions handled in the same way. 

It is assumed that the goal is the state at the end of the demonstration. Demonstrations with the same goal conditions are grouped together and a BT is generated for each group. Finally, the planner combines these trees under a fallback node. In this way, if the task represented by one subtree fails to achieve the goal, another subtree is executed to attempt achieving the same goal.

\subsection{Disambiguation of Requests}
\label{disambiguation}

\begin{figure*}[tbp]
    \centering
    \includegraphics[width=1\linewidth]{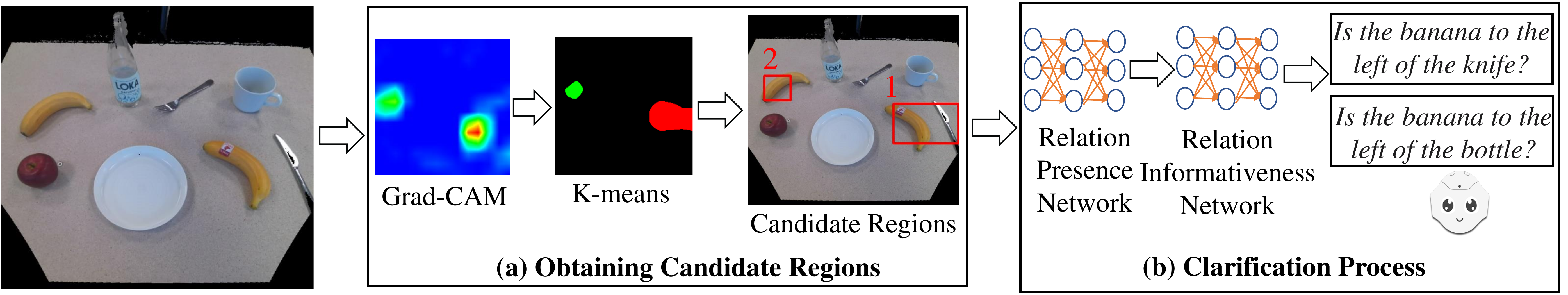}
    \caption{The disambiguation process when the robot aims to find the `banana'.}
    \label{fig:disambiguation}
\end{figure*}
%

In the cases where the task fails and \textit{`Scene Clear?'} condition turns \texttt{False}, we employ the method from~\cite{dogan_followup_2022} that asks for follow-up clarifications to resolve ambiguities. This method obtains the candidate regions that are causing the ambiguities leveraging explainability~\cite{dogan2021leveraging} and generates yes/no questions for each of the candidates describing them uniquely with their spatial relations~\cite{8968510}. Once the system receives an affirmative answer to a question for any of the candidates, this candidate is suggested as the one described by the user -- see Figure~\ref{fig:disambiguation} for an overview.

For a given RGB scene and an expression describing an object in the scene, an activation heatmap $H$ showing the areas contributing to the predictions are obtained using the image captioning module of Grad-CAM~\cite{selvaraju2017grad}. Then, the generated heatmap $H$ is clustered with K-means clustering. The number of clusters is initialized with the number of unconnected regions in $H$ and determined using the 2D connectivity of the pixels -- i.e., if 2 pixels are active and neighbors of each other in 2D, they are assumed to be connected. After obtaining the clusters, they are sorted according to their corresponding activations in $H$, and the sorted clusters become the set of candidate regions $C$ (Figure~\ref{fig:disambiguation}a).

To generate the follow-up questions uniquely referring to each candidate in $C$, the set of objects $O$ in the scene are detected using the DETR object detector~\cite{carion2020end} (if the target object category is detectable, $C$ is reassigned with the objects in $O$ that are closest to candidate regions and belong to the target object category). Then, the spatial relations between each candidate in $C$ and each object in $O$ are determined using the Relation Presence Network (RPN)~\cite{8968510}. This network takes a pair of objects and outputs the spatial relationship between them, such as `to the left', `to the right', `in front', `behind', or `close to'. After finding the spatial relations, we use the Relation Informativeness Network (RIN)~\cite{8968510} to find the most informative relation that describes each candidate object. Similar to RPN, this network takes a pair of objects and the spatial relations between them and outputs an informativeness value to refer to objects unambiguously.

After finding the most informative spatial relation describing each candidate in $C$, a yes/no question is formed by spatially describing these candidates (Figure~\ref{fig:disambiguation}b). e.g, when the request is `the white plate', the clarification question is formed as `Is the white plate to the left of the banana?'. If an affirmative answer is received for any of the candidates, this candidate is suggested.


\section{Experiments and Results} \label{sec:experiments}


In the following section we perform experiments to highlight the benefits of being able to disambiguate a task. All experiments are performed with the same steps: the BT is learned by performing the task three times in non-ambiguous scenarios, then ambiguities can arise during the execution. We perform tests with increasing levels of ambiguity, where the task to perform is always the same: to pick a banana and drop it in a bowl.

The experiments are performed using an ABB YuMi robot with an Azure Kinect camera mounted on top of it. The objects used for the experiments are chosen from those represented in the COCO dataset\footnote{\url{https://cocodataset.org}}. We capture the users' request using the Google speech recognition engine~\cite{google}, and converted it to the text, then the disambiguation questions are output through a Jabra speaker. The detection algorithm and the disambiguation framework networks run in a computer with 3 NVidia GTX 1080 GPUs. The computer OS is Linux Ubuntu 20.04 with ROS2 Foxy.
The BT is ticked at a frequency of $0.4$~Hz and the disambiguation framework takes approximately $30$~seconds for a full iteration.

\paragraph{\textbf{BT without disambiguation}}
In this experiment we do not include the disambiguation subtree in the BT. If the scene during execution is unambiguous, i.e. every object type is only present once, then the robot is able to perform the task without external intervention from the user. If the scene is ambiguous, e.g. there are two bananas, the robot is not able to complete the task as it is not able to evaluate the condition \emph{`Object roughly At?'} in Figure~\ref{fig:step1}, nor perform the pick behavior, as there would be frames called \textit{banana\_1} and \textit{banana\_2}, but not \textit{banana}. The robot is not able to recover from this situation, unless a user removes one of the two ambiguous objects from the scene. Note that, although possible, making the robot pick one of the objects at random would not be a satisfactory solution of the task.

\paragraph{\textbf{BT with disambiguation, execution not ambiguous}}
This test is logically equivalent to the previous one in the unambiguous case. The final BT would have just three more nodes, but since the \emph{`Scene Clear?'} condition is initialized as \texttt{True}, the disambiguation behavior is never executed if the task is not ambiguous.

\paragraph{\textbf{BT with disambiguation, ambiguous pick target}}\label{expC}
In this experiments, we insert a second banana in the scene at execution time. As for the ambiguous case in experiment \textit{a)}, the robot fails in evaluating the condition. In this case however, the disambiguation behavior is triggered and the robot can ask the user follow-up questions to disambiguate the scene. The robot then picks whichever object the user indicates as the target for the task.

\paragraph{\textbf{BT with disambiguation, ambiguous place pose}}
In this case, we have an unambiguous target for the \emph{`Pick'} action, but we provide two bowls. So the reference frame for the \emph{`Drop'} action and the \emph{`Object Roughly At?'} condition is ambiguous. The disambiguation framework is triggered and user intervention is required, and then the robot is able to successfully complete the task.

\paragraph{\textbf{BT with disambiguation, ambiguous pick and place}}
This experiment combines the two previous ones. The steps in the execution process are reported in Figure~\ref{fig:experimentE}. The key for a correct execution is in the \emph{`Object Roughly At?'} condition. The condition computes all the transformations for the following manipulation (pick then place) subtree. First the transformation from the reference frame of the \emph{`Drop'} action to the robot base frame is computed, which triggers the disambiguation framework with query \textit{`bowl'}. Then, the transformation from the target item for the \emph{`Pick'} action to the robot base frame is computed, triggering the disambiguation framework with query \textit{`banana'}. If instead the opposite would happen, we would have a situation where the robot attempts picking the \textit{banana} but then it stops because of the reference frame \textit{bowl} not being available. This behavior would create unnecessary noise in the detection algorithm, jeopardizing the success of the task. The operator could also exploit the reactivity of the system and remove a banana from the scene while disambiguating the bowl. In this case, the frame names update automatically, and since there is now just one banana, execution continues without running the disambiguation behavior again\footnote{Also showed in the video at \url{https://youtu.be/aC1wY35ZNWk}}.

For sanity check, to ensure the robot could resolve the task, we ran the experiment with a single banana and a bowl without our disambiguation component, and the case was a success. Also, as a proof of concept with the disambiguation component, we asked four different ABB employees that were not familiar with the system to test it with the case scenario \ref{expC}, from demonstration to execution. The experiment was successful in 3 out of 4 cases, and the unsuccessful case happened because of the object detection failure. Although further experiments can be conducted for detailed analysis, this case study shows that our system achieves to learn the task from demonstration and is capable of disambiguation during execution.

\section{Conclusions and Future Work} \label{sec:conclusion}

In this paper we have presented a system that integrates previous work that learns Behavior Trees from demonstration, with a framework to disambiguate objects in the scene. The system allows an operator to show a task to the robot in a non ambiguous setting. The task is learned by the robot in the form of a BT and a subtree to run the disambiguation framework is automatically attached. With this system, during the task execution the robot is able to request human help, by asking follow-up questions with the goal of finding the target object for the task, disambiguating it. The system is tested in a manipulation task.

Due to the restrictive measures enforced by ABB to face the COVID-19 pandemic, limiting the number of people allowed in the laboratory facilities, it was not possible to conduct a similar user study as in \cite{dogan_followup_2022}. Such a study is left as future work, where we plan on extending the disambiguation method also to the demonstration and learning part. 
A more robust perception system, integrating views from different poses to reduce the obstruction caused by the robot arms, would increase the overall robustness of the system. 
Another beneficial extension would be to endow the system with the possibility of using both arms and different grasping tools (parallel grippers or suction cups) during the demonstration, so to record the used arm and grasping tool alongside the performed actions. In such a case the working range of the robot would increase and it would be possible to demonstrate more complicated tasks. Note that there is a limited number of tasks that are possible to perform with objects from the COCO dataset that are easy and small enough to grasp. The system would benefit from a dedicated dataset with industrial tools, so it would be possible to perform tasks such as kitting and stacking (similar to what shown in~\cite{gustavsson_combining_2021}). In such case, all neural networks would have to be re-trained.

\section*{Acknowledgements}
Authors would like to thank Leonard Bruns from the Division of Robotics, Perception and Learning, KTH, for the precious support provided in designing the perception layer.

\bibliographystyle{IEEEtran}
\bibliography{references}

\end{document}